\documentclass{article}
\usepackage{ijcai20}

\usepackage{times}
\usepackage{soul}
\usepackage{url}
\usepackage[hidelinks]{hyperref}
\usepackage[utf8]{inputenc}
\usepackage[small]{caption}
\usepackage{graphicx}
\usepackage{amsmath}
\usepackage{amsthm}
\usepackage{booktabs}
\usepackage{algorithm}
\usepackage{algorithmic}
\usepackage{epsfig}
\usepackage{subfigure}
\usepackage{array}
\usepackage{indentfirst}
\usepackage{pifont}
\usepackage{multirow}
\usepackage{color}
\usepackage{graphicx}
\usepackage{tikz}
\usetikzlibrary{positioning}
\title{When Residual Learning Meets Dense Aggregation: Rethinking the Aggregation of Deep Neural Networks}

\author{
    Paper ID: 4194
}

\author{
Zhiyu Zhu$^{1}$
\and
Zhen-Peng Bian$^{2}$\and
Junhui Hou$^{1,*}$\and
Yi Wang$^3$\And
Lap-Pui Chau$^3$
\affiliations
$^1$City University of Hong
Kong\\
$^2$Singapore Telecommunications Limited\\
$^3$Nanyang Technological University\\
\emails
zhiyuzhu2-c@my.cityu.edu.hk,
zbian1@ntu.edu.sg,
Yi Wang@e.ntu.edu.sg,
jh.hou@cityu.edu.hk,
elpchau@ntu.edu.sg
}
\begin{document}

\maketitle

\begin{abstract}
	Various architectures (such as GoogLeNets, ResNets, and DenseNets) have been proposed. However, the existing networks usually suffer from either redundancy of convolutional layers or insufficient utilization of parameters. To handle these challenging issues, we propose Micro-Dense Nets, a novel architecture with global residual learning and local micro-dense aggregations. Specifically, residual learning aims to efficiently retrieve features from different convolutional blocks, while the micro-dense aggregation is able to enhance each block and avoid redundancy of convolutional layers by lessening residual aggregations. Moreover, the proposed micro-dense architecture has two characteristics: pyramidal multi-level feature learning which can widen the deeper layer in a block progressively, and dimension cardinality adaptive convolution which can balance each layer using linearly increasing dimension cardinality. The experimental results over three datasets (i.e., CIFAR-10, CIFAR-100, and ImageNet-1K) demonstrate that the proposed Micro-Dense Net with only 4M parameters can achieve higher classification accuracy than state-of-the-art networks, while being 12.1$\times$ smaller depends on the number of parameters. In addition, our micro-dense block can be integrated with neural architecture search based models to boost their performance, validating the advantage  of our architecture. We believe our design and findings will be beneficial to the DNN community.
\end{abstract}
\section{Introduction}
\setlength{\parindent}{2em}
Owing to the great representation ability, deep neural networks (DNNs) have achieved great success in machine learning and computer vision communities \cite{gu2018recent,rawat2017deep}.
\footnote{Z.Zhu and Z.-P. Bian contributed to this work equally. \textsl{Corresponding author: Junhui Hou}}{
	\begin{figure}[h]
		\setlength{\belowcaptionskip}{-0.5cm}
		\centering
		\includegraphics[width=1\linewidth]{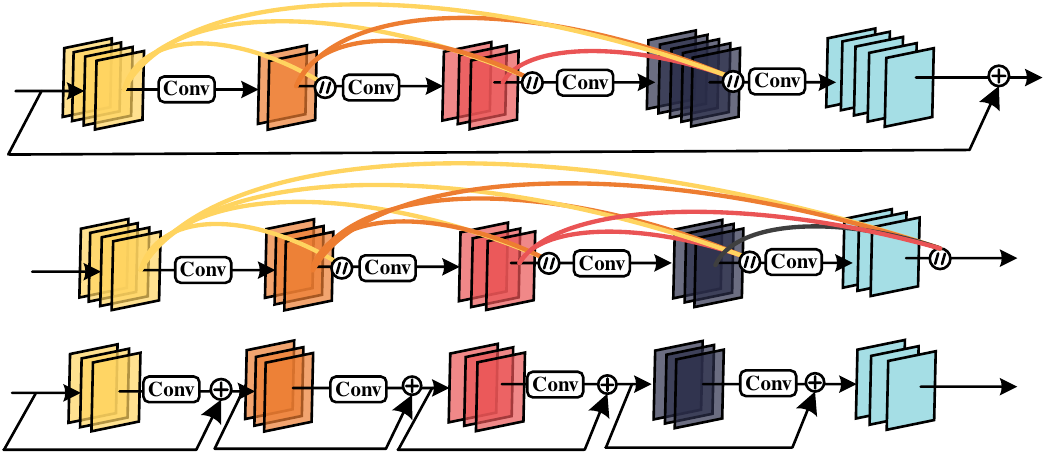}
		\caption{Qualitative comparisons of the proposed network architecture with the well-known DenseNets and ResNets. \textsl{\textbf{Top}}: Our architecture, miro-dense aggregation with global residual learning. \textsl{\textbf{Middle}}: Dense aggregation of	DenseNets, \textsl{\textbf{Bottom}}: residual learning of ResNets. Our proposed network architecture uses local dense and global identity mapping aggregations. Moreover, to make full use of parameters, the proposed micro-dense structure has two characteristics: (1) \textbf{pyramidal multi-level feature learning}, i.e., gradually increase the output channel of densely connected layers; and (2) \textbf{dimension cardinality adaptive convolution}, i.e., rise the groups of convolutional layers with the growth of width. '+' denotes the addition operation and '//' denotes the concatenation operation.}
		\label{Micro-Dense}
	\end{figure}
	\begin{figure*}[htbp]
		\setlength{\abovecaptionskip}{-0.4cm}
		\setlength{\belowcaptionskip}{-0.6cm}
		\centering
		\includegraphics[width=\textwidth]{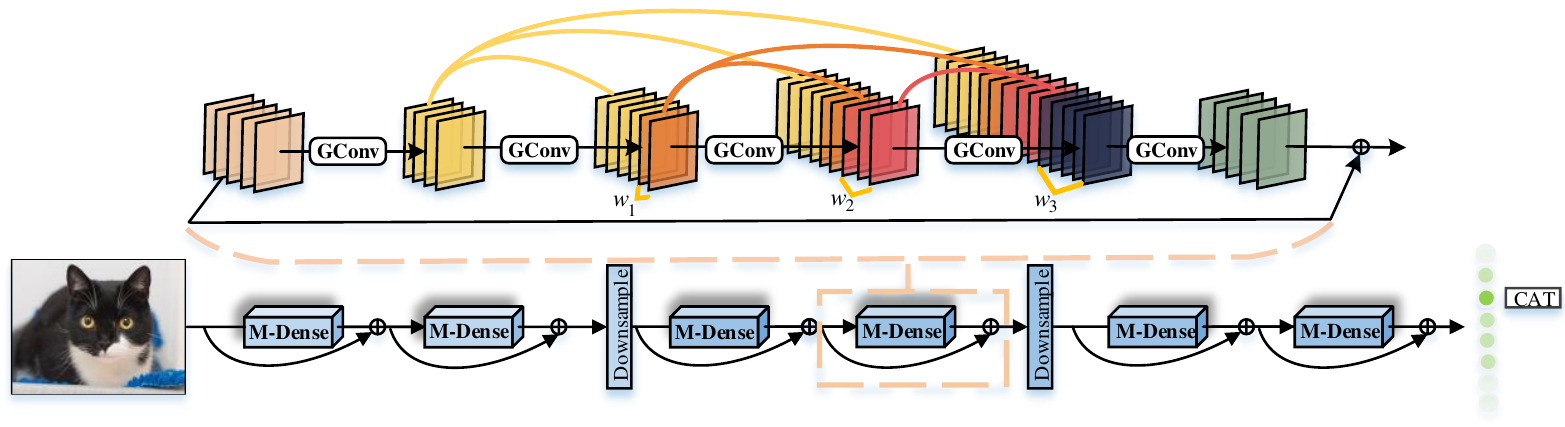}
		\label{arch_all1}
		\caption{Illustration of our Micro-Dense Nets and micro-dense aggregations (noted as M-Dense in the figure), where $w_{1}<w_{2}<w_{3}$ indicates that both the output channels and dimension cardinalities are increasing while convolutional layer is going deeper, and GConv represents a block of GroupedConvolution-BatchNormalization-ReLU.}
\end{figure*} 
Since Alexnet \cite{krizhevsky2012imagenet}, a plain network that contains stacked convolutional layers, design of DNNs has become a hot research topic. Nowadays, the neural network architecture is going towards deeper and deeper \cite{simonyan2014very}. However, as the depth of DNNs scales up, problems such as gradient vanishing and overfitting appear and make them difficult to train \cite{Glorot2010Understanding}. To address these problems, DNNs with different topologies have been proposed, e.g., highway networks \cite{srivastava2015highway} with shortcuts followed by ResNets \cite{he2016deep} using identity mappings and DenseNets \cite{huang2017densely} with dense concatenation. Meanwhile, some training techniques e.g., batch/group normalization \cite{ioffe2015batch,wu2018group} and drop-out layers \cite{srivastava2014dropout}, were also proposed to deal with the optimization of DNNs. Although training very deep neural networks becomes easier with the help of shortcuts and training techniques, the performance of DNNs does not gain much from that extreme depth. \cite{he2016deep}\\
\indent Instead of increasing the depth, GoogLeNets \cite{szegedy2015going} enlarge the width (i.e., the number of channels) of the DNN by using parallel convolutional layers with different kernel sizes. ResNeXts \cite{xie2017aggregated} try to use cardinality to extract deep features efficiently. And Res2Nets \cite{gao2019res2net} use features from different levels to augment the extraction ability. All of them share the same conception of using multiple branches rather than single convolutional layers in residual learning.\\
\indent DenseNets \cite{huang2017densely} reveal that the redundancy of convolutional layers might be the reason why the improvement stops while ResNets go deeper. Hence, the dense aggregation was proposed to handle the problem. The importance of feature reuse was also demonstrated. However, dense aggregation consumes plenty of resources on extracting accumulated features from all proceeding layers. SparseNets \cite{zhu2018sparsely} presents a sparse connection to reduce the heavy burden of the dense aggregation. Moreover, some works try to fuse the ResNets and DenseNets together, e.g., Dualpath networks \cite{chen2017dual}, residual dense blocks \cite{zhang2018residual}, and Mixed Link Networks (MixNets) \cite{wang2018mixed}. All these methods aim to make full use of residual learning and dense aggregation.\\
\indent \textbf{Our Solution.} In this paper, by comprehensively analyzing the different types of aggregations in existing DNNs, we propose Micro-Dense Nets, which  learns features in locally dense and globally residual manners. As shown in Figure 2, Micro-Dense Nets consist of several sub-convolutional blocks named micro-dense blocks, which densely connects local layers inside the block to extract features. Meanwhile, identity mapping is adopted to connect both ends of the micro-dense block. It inherits the advantages of multi-level feature learning and efficient feature storage from dense aggregations and residual learning, respectively. Moreover, the proposed micro-dense block possesses the following two characteristics: 1) pyramidal multi-level feature learning, i.e., subsequent layers are wider than preceding ones in a micro-dense block to avoid the loss of information; and 2) dimension cardinality adaptive convolution, i.e., in the proposed micro-dense block, the number of groups of convolutional layers gradually increases with the dimension of feature-maps to avoid parameter explosion, which will be aroused by the pyramidal multi-level feature learning strategy. \\
\indent \textbf{Our Findings.} Through comprehensive experiments on ImageNet-1K and CIFAR, the proposed Micro-Dense Nets exceed state-of-the-art methods and reach 97.84\% and 84.64\% on CIFAR-10 and CIFAR-100, while being at most 12.1 $\times$ smaller than state-of-the-art DNNs. Moreover, we integrate our Micro-Dense Nets with the state-of-the-art neural architecture search (NAS) based model and obtain 0.5\% gain in terms of the Top 1. classification accuracy on ImageNet-1K, which validates the advantage of our architecture. Last but not least, based on the experimental results, we can draw the following conclusions: \textsl{1) dense aggregation indeed boosts the feature extraction ability of DNNs, but very deep dense aggregation may degrade the performance of DNNs with the same number of parameters; and 2) although conventional multi-level features with a fixed growth rate are beneficial to DNNs, the multi-level feature learning with a varying growth rate, i.e., our pyramidal multi-level feature learning strategy, will be a better choice. Accordingly, a well designed method for avoiding parameter explosion should accompany. We believe our design and findings will be beneficial to the DNN community.}  \\
% pyramidal multi-level feature learning indeed further boost the performance of DNNs.\\
\section{Revisiting Aggregated DNNs}
\indent Aggregations play a critical role in DNNs. They could strengthen the feature extraction ability of DNNs, e.g., shortcuts existed in ResNets and dense aggregations in DenseNets. However, inappropriate introduction of aggregations may be harmful to DNNs. In the following, we will analyze the four types of aggregations adopted by existing DNNs comprehensively.
\subsection{Aggregations in DNNs}
\indent Aggregations adopted in existing DNNs can be roughly classified into four categories, i.e., plain, highway connection, dense aggregation and inception. To use both features and parameters efficiently, combining diverse aggregations in multiple levels is a promising solution. We discuss such a combination with detailed analysis. In the following, we denote $\mathcal{H}(\cdot)$ as the composite function of operations including Convolution (Conv), Pooling, Batch Normalization (BN), and Recitified Linear Units (ReLU).\\
\indent \textbf{Plain}. As a basic connection in DNNs, the plain aggregation means to simply stack different layers together. Therefore the output can be represented as
\begin{equation}
x^{k}=\mathcal{H}_{k-1}(x^{k-1}),
\end{equation}
where $x^{k}$ mean the output feature-maps of $k$-th layer. Limited by the simple topology, plain connections have a weak feature learning ability. Considering the fact that the topology of DNNs helps to enhance representation learning \cite{chen2017dual}, such a simple plain connection may even degrade the feature learning ability of networks.\\
\indent \textbf{Highway connection}. The hghway connection \cite{srivastava2015highway} expressed as:
\begin{equation}
\begin{split}
\resizebox{0.88\linewidth}{!}{$
x^{k}= x^{k-1}\times R_{k-1}(x^{k-1}) + \mathcal{H}_{k-1}(x^{k-1}) \times T_{k-1}(x^{k-1}),
$}
\end{split}
\end{equation}
where $R_{k-1}$ and $T_{k-1}$ are two nonlinear transforms named carry gate and transform gate, respectively. ResNets adopt identity mapping where $R_{k-1}(x^{k-1})$ and $T_{k-1}(x^{k-1})$ are set to an identity function leading to:
\begin{equation}
\begin{split}
x^{k}= x^{k-1} + \mathcal{H}_{k-1}(x^{k-1}).
\end{split}
\label{reseq}
\end{equation}
In residual learning, each layer is fed by the sum of multiple output feature-maps of preceding layers. Different from the plain network, identity mapping in ResNets could not only help the feature propagation but also alleviate the gradient vanishing problem.\\
% However, frequently feature additions and weak learning abilities of residual blocks may cause redundancy of convolutional layers.\\
\indent \textbf{Dense aggregation}. Dense aggregation preserves all the output feature-maps in preceding layers. And the output of dense aggregation is outlined as:
\begin{equation}
\begin{split}
x^{k}= \mathcal{H}_{k-1}([x^{1},x^{2},...,x^{k-1}]),
\end{split}
\end{equation}
where $[x^{1},x^{2},...,x^{k-1}]$ refers to the concatenation of the feature-maps produced by preceding $1,2,...,k-1$ layers. Rather than the hard-code identity mapping in residual learning, dense aggregation learns to weighted-sum different level features. Due to the flexibility of multi-level feature learning, dense aggregations boost the feature learning ability of DNNs. Nonetheless, accumulated feature-maps are heavy burdens of Dense-aggregations which restrict effective parameter using.\\
\indent \textbf{Inception}. The inception block uses several plain convolutional blocks to extract features separately and then concatenates them. Inception aggregation can be expressed as:
\begin{equation}
\begin{split}
x^{k} = [\mathcal{H}_{k-1}^{1}(x^{k-1}),\mathcal{H}_{k-1}^{2}(x^{k-1}),...,\mathcal{H}_{k-1}^{j}(x^{k-1})].
\end{split}
\end{equation}
Inception aggregations widen DNNs and could also combine features in different levels. However, in deep inception aggregation, each branch extract features separately, and convolutional layers could only percepte feature in one branch. Thus, deepening inception aggregation reduces the richness of features and may degenerate the performance of DNNs.
\subsection{Assembling of Different Aggregations}
\indent Since network in network \cite{lin2013network} was proposed, the research on combining
different kinds of aggregation in DNNs has never stopped.
\cite{wang2018mixed} found that both ResNets and
DenseNets come from the same dense topology. However,
they varied in the way of aggregations. DualpathNets \cite{chen2017dual} have both addition and concatenation
connections, and it could save the different level features and
realize residual feature learning.\\
\indent Identity mapping in residual learning greatly boosts the feature learning ability of DNNs. However, as the residual learning based networks get deeper, frequently residual feature additions cause the redundancy of convolutional layers \cite{huang2017densely}, which will degrade its performance. Res2Nets indicate that instead of increasing the number of residual blocks in residual learning-based architectures, enhancing feature extraction ability of each block can be a promising solution.
%Through increasing the depth of the residual block, DNNs could achieve the same depth but with fewer residual aggregations, which may boost the feature extraction ability of each block and alleviate the redundancy of convolutional layers.\\
%\indent 
However, the convolutional layers of Res2Nets only reuse features from two levels, leading to a weaker representation ability when compared with the dense aggregation in DenseNets which could utilize features in multiple levels. Unfortunately, the dense aggregations occupy too many resources for feature-maps storage and fusion. Also widening or deepening a dense block will cause high resource consumption \cite{zhu2018sparsely}. Additionally, a critical point neglected by most dense aggregations based methods \cite{huang2017densely,zhu2018sparsely,chen2017dual,wang2018mixed} is that all the dense aggregation based methods adopt a fixed channel growth rate for dense layers. Due to feature accumulation, the number of input channels, which restricts the maximum information, increases while the dense layer deepens. Thus, the fixed rate may result in losses of information. Notwithstanding this drawback, intuitively increasing the growth ratio on layer-wise is impossible for dense aggregation based methods. The reason is that it will lead to both features and parameter explosion. That is, the number of input channels (the number of parameters) grows from $C_{n}$ to $\hat{C}_{n}$:
\begin{equation}
\vspace{-1em}{
C_{n}=c_{0}+\sum_{l=1}^{n}c_{i} =  c_{0}+ n\times r_{0},}
\label{channel_2}
\end{equation}
\begin{equation}
\vspace{-1.6em}{
	c_{i} = r_{0}}
\end{equation}
\begin{equation}
\vspace{-1.2em}{
	\resizebox{0.88\linewidth}{!}{$
\hat{C_{n}} =c_{0}+\sum_{l=1}^{n}\hat{c_{i}}= c_{0}+n\times r_{0} + \hat{r}\times (n-1)\times n/2,$}
}
\label{channel_1}
\end{equation}
\begin{equation}
\vspace{0em}{
\hat{c_{i}} = r_{0} + \hat{r}\times (i-1),}
\end{equation}
where $C_{n}$ is the channel number of the fixed growth-rate $r_{0}$ and $n$ preceding layers and $\hat{C_{n}}$ represents the channel number with the growth rate gradually increasing ($\hat{r}$ is increasing factor of growth rate $r_{0}$).\\
\indent \textbf{Remark.} From the above analysis, we can conclude that: 1) there are only two operators for feature fusion: addition ('$+$' represented by residual learning) and concatenation ('$[..]$' represented by DenseNets). The former one could efficiently preserve features from multiple levels without consuming extra space, while the latter one reuses all output features from preceding layers to strengthen the richness of features; and 2) although multi-level feature reuse could enhance the feature learning ability of DNNs, high level features extracted by multiple layers have stronger representation abilities. In order to enhance feature extraction ability of DNNs, we should scale them up. However, ResNets are limited by the redundancy of convolutional layers and DenseNets consume too much on feature-maps reduction. Meanwhile, DenseNets fix the number of output channels which restricts the high level features.\textsl{Based on these analyses, we propose a new architecture , which can inherit the advantages of the existing DNNs well and avoid the their drawbacks such that performance is boosted.}
\begin{table}
	\scriptsize
	\setlength{\belowcaptionskip}{-0.5cm}
	\begin{tabular}{m{0.10\columnwidth}<{\centering} m{0.1\columnwidth}<{\centering} m{0.21\columnwidth}<{\centering} m{0.25\columnwidth}<{\centering} m{0.10\columnwidth}<{\centering}}
		\toprule
		\multicolumn{2}{c}{Layer}					&Input channel $c_{i}$  		&Output channel $c_{u}$		 			& Cardinality \\
		\hline
		\multirow{2}*{Compression} &BatchNorm 	&$c_{i}$						&$c_{i}$									&-\\
		&C1-BR		&$c_{i}$						&$c_{0}= \lfloor c_{i}\times r_{a} \rfloor \times g_{c}$&-\\
		\hline
		\multirow{2}*{Dense-$1$}   &C1-BR	 	&$\hat{c}_{1}=c_{0}$			&$c_{1}= c_{0} / g_{c} \times k_{1}$				&-\\
		&C3-BR		&$c_{1}$						&$c_{1}$									&$k_{1}$\\
		\hline
		\multirow{2}*{Dense-$2$}   &C1-BR	 	&$\hat{c}_{2}=c_{0}+c_{1}$		&$c_{2}= c_{0} / g_{c} \times k_{2}$					&-\\
		&C3-BR		&$c_{2}$						&$c_{2}$									&$k_{2}$\\
		\hline
		\multirow{2}*{Dense-$3$}   &C1-BR	 	&$\hat{c}_{3}=c_{0}+c_{1}+c_{2}$&$c_{3}= c_{0} / g_{c} \times k_{3}$					&-\\
		&C3-BR		&$c_{3}$						&$c_{3}$									&$k_{3}$\\
		\hline
		%		\vdots		&\vdots	 	&\vdots							&\vdots							&\vdots\\
		\multirow{2}*{Dense-$n$}  &C1-BR	 	&$\hat{c}_{n}=\sum_{i=0}^{n-1}c_{i}$ & $c_{n}= c_{n} / g_{c} \times k_{n}$			&-\\
		&C3-BR		&$c_{n}$						 &$c_{n}$									&$k_{n}$\\
		\midrule
%		Output Layer 			  &C1-BR	 	&$c_{n+1}=c_{1}+c_{2}+..+c_{n}$&$c_{u}$								&-\\
		Output Layer 			  &C1-BR	 	&$c_{n+1}=\sum_{i=0}^{n}c_{i}$&$c_{u}$								&-\\
		\bottomrule
	\end{tabular}
	\vspace{-1em}
	\caption{The detailed architecture of a micro-dense block with $n$ dense layers. C$s$-BR indicates a stack of Convolution-BatchNorm-Relu layer with kernel of size $s\times s$.  Cardinality is the number of groups and $g_{c}$ is set to be 4. The micro-dense block with $n$ dense layers is denoted as \textsl{MDConv-n}.}
	\label{micro-dense archcitecture}
\end{table}
\section{The Proposed Micro-Dense Nets}
As shown in the bottom of Figure 2,  our Micro-Dense Nets consist of multiple convolutional blocks named micro-dense block, which uses dense aggregations to link local layers shown in the top of Figure 2. Meanwhile, in order to strengthen feature propagation, we use identity mapping to connect both ends of a micro-dense block. In what follows, more details will be provided about the proposed Micro-Dense Nets.
\subsection{Local Micro-Dense Architecture}
The proposed micro-dense block is constructed by several sub-bottleneck structures. Different from the conventional dense layer, it has following two characteristics:\\
\textbf{Pyramidal multi-level feature learning}. In each micro-dense block, convolutional layers learn multi-level features with the output feature dimension gradually increasing. Although the conventional dense aggregation \cite{huang2017densely} with a fixed growth rate boosts the performance of DNNs, the inconsistency between the input/output feature dimensions may degrade its representation ability. Moreover, to the best of our knowledge, all dense aggregations based architectures \cite{zhu2018sparsely,chen2017dual,wang2018mixed,zhang2018residual} took it for granted that the number of output channels is fixed. In the proposed micro-dense block, this problem could be relieved, as each micro-dense block is quite shallower than the full-dense aggregation. Similar to \cite{han2017deep}, we increase the width of dense layers inside a dense block linearly, i.e.,
\begin{equation}
c_{l} = (l+1)\times c_{0},
\label{MDenseIncre}
\end{equation}
where $c_{l}$ is the width of the $l$-th dense layer in a micro-dense block. Aggregating the dense layers locally inside a micro-block is able to relieve the feature explosion. Because dense aggregation only has to preserve features in one block. Nevertheless, such a linear manner also leads to parameter explosion. Since the representation learning ability of a single convolutional layer is controlled by the number of parameters, each layer should have a similar number of parameters inside a block. To this end, we introduce dimension cardinality adaptive convolution to achieve efficient parameter utilization.\\
\textbf{Dimension cardinality adaptive convolution}. Dimension cardinality indicates the number of groups in convolutional layers. Unlike the conventional convolution, it divides the input into several groups and convolves with several shallow kernels separately on these feature-maps. Thus, it could reduce the number of parameters in a single layer. As shown in Table \ref{micro-dense archcitecture}, the number of groups grows up while the convolutional layers are deepening, which will greatly reduce the number of parameters. Specifically, the process is expressed as:
\begin{equation}
\vspace{-1.6em}{
P_{l} = c^{u}_{l}\times(c^{i}_{l}\times w_{l}\times h_{l}+1),}
\label{normal_p}
\end{equation}
\begin{equation}
\vspace{0em}{
\resizebox{0.88\linewidth}{!}{$
P^{g}_{l} = G_{l}\times(c^{i}_{l}/G_{l})\times((c^{u}_{l}/G_{l})\times w_{l}\times h_{l}+1)\approx P_{l}/G_{l},
$}}
\label{group_p}
\end{equation}
where ($P_{l}^{g}$ / $P_{l}$) indicates the number of parameters in the convolutional layer (with / without) groups, $c_{i}$ and $c_{u}$ are the input and output channels, and $w_{k}$ and ${h}_{k}$ represent the width and height of the corresponding kernel. According to Eq.\ref{normal_p}, if the channels $c_{i}$ and $c_{u}$ grow linearly with respect to $l$ ($l$ is the depth of dense layers shown in Eq.\ref{MDenseIncre}), the number of parameters grows quadratically with respect to $l^{2}$. By increasing $G^{k}$ linearly as $\beta \times l$, we could alleviate the problem of parameter explosion (from quardractic to linear). In a micro-dense block, the cardinality grows linearly with
\begin{equation}
\vspace{0em}{
		k_{n} = n+1.
		}
\end{equation}
\indent Furthermore, the group convolution requires both input and output channels to be dvisible by the group number. Thus, we design the bottleneck structure for each layer to round the channel of input feature-maps. In summary, Table \ref{micro-dense archcitecture} lists the detailed architecture of our micro-dense block.
\subsection{Global Residual Learning}
In order to achieve storage of efficient feature-maps , we use identity mappings to connect different blocks of neural network globally. Till now, various residual learning based methods have been proposed e.g., ResNets \cite{he2016deep} and PyramidNets \cite{han2017deep}. Due to the gradual increment of feature-maps, PyramidNets have good performance compared to other residual learning-based methods. Thus, we use the linear increasing for pyramidal residual learning \cite{han2017deep} as global aggregation. Meanwhile, the number of feature-map channels increases as:
\begin{equation}
\resizebox{0.8\linewidth}{!}{$
\mathcal{W}_{k}^{\alpha}=\lfloor \mathcal{W}_{0} + k\times\alpha / N \rfloor, \quad if\quad 0\leq k \leq N,$}
\label{pyramid}
\end{equation}
where $N$ denotes the total number of micro-dense blocks, $\alpha$ is the additional channels in Micro-Dense Nets, $\mathcal{W}_{k}^{\alpha}$ is the output channel number of the $k$-th micro-dense block, which is increased by a step factor of $\alpha/N$, and $\mathcal{W}_{0}$ is the number of input channels to first block which is set it to 16 according to \cite{han2017deep}.
\section{Experiments and Discussion}
	\indent We evaluate the proposed Micro-Dense Nets on several commonly-used benchmark datasets and compare with state-of-the-art DNNs, especially residual learning and dense aggregation based methods. We also compare with state-of-the-art NAS based methods, e.g., FPNASNet \cite{cui2019fast}. %Proxyless-G \cite{cai2018proxylessnas}.
	\begin{table}
		\scriptsize
		\setlength{\belowcaptionskip}{-0.4cm}
		\resizebox{\linewidth}{!}{
		\begin{tabular}{m{0.12\columnwidth}<{\centering} m{0.1\columnwidth}<{\centering} m{0.29\columnwidth}<{\centering} m{0.32\columnwidth}<{\centering} }
			%		\begin{tabular}{c c c c}
			\toprule
			Stage					& input					&Micro-Dense Net-30$\alpha $64  			&Micro-Dense Net-60$\alpha $115 	  \\
			\hline
			Conv1 					&32$\times$32 				&C3-B								&C3-B									\\
			\multirow{2}*{Conv2}   	&\multirow{2}*{32$\times$32}&\multirow{2}*{[$MDConv-3$]$\times$10}&\multirow{2}*{[$MDConv-3$]$\times$20}						\\
			&							&		\\
%			Downsample1				&16$\times$16				&$MDConv-s2$								&$MDConv-s2$					\\
			\multirow{2}*{Conv3}	&\multirow{2}*{16$\times$16}&\multirow{2}*{[$MDConv-3$]$\times$10}	&\multirow{2}*{[$MDConv-3$]$\times$20}						\\
			&\\
%			Downsample2				&8$\times$8					&$MDConv-s2$								&$MDConv-s2$							\\
			\multirow{2}*{Conv4}	&\multirow{2}*{8$\times$8}	&\multirow{2}*{[$MDConv-3$]$\times$10}	&\multirow{2}*{[$MDConv-3$]$\times$20}				\\		
			&\\
			\multirow{2}*{Output}	&1$\times$1					&\multicolumn{2}{c}{8$\times$8 global average pool}				\\
			&10 / 100					&\multicolumn{2}{c}{ full connected layer, soft max}	\\
			\bottomrule
		\end{tabular}}
		\vspace*{-0.8em}
%		\caption{The detailed architecture of Micro-Dense Nets for CIFAR classification. $MDConv-s2$ indicates the micro-dense block using one dense layer with the kernels of size $4\times4$ and stride $2$. }
		\caption{The detailed architecture of Micro-Dense Nets for CIFAR classification. The first block of Conv3, Conv4 use one conolutional layer of kernel of size 4 $\times$ 4 and stride of 2 to downsample the feature-maps. $\alpha$ is the width factor shown in Eq. \protect\ref{pyramid}.}
		\label{cifar_arch}
	\end{table}
\subsection{Datasets and Implementation Details}
The experiments were carried on two datasets: CIFAR dataset \cite{krizhevsky2009learning} and ILSVRC 2012 classification dataset (ImageNet-1K) \cite{deng2009imagenet}. Specifically, CIFAR contains colored nature images of resolution 32$\times$ 32 with two sub-datasets: CIFAR-10 from 10 classes object images and CIFAR-100 from 100 classes. The training and test sets of both CIFAR-10 and CIFAR-100 are 50,000 and 10,000 images, respectively. We report the final
test error at the end of training on the 10K images from the test set. The ILSVRC 2012 classification dataset (ImageNet-1K) contains 1,000 classes images with 1.2 million images for training, and 50,000 for validation. The classification error was calculated on the validation set. We used auto-augmentation \cite{cubuk2019autoaugment} to train neural networks. \\
\indent All the networks were trained using stochastic gradient descent(SGD). On CIFAR, the batchsize being 128 For ImageNet-1K, we train neural network with batchsize 400. The learning rate was initialized to 0, and gradually changed as:
\begin{equation}
\vspace{-0.8em}{
\resizebox{.85\linewidth}{!}{$
	lr= \begin{array}{ll}
	lr_{top}\times(1+cos(\pi\times i/N_{a} ))/2 & \forall i \in [0,N_{a}]
	\end{array}$,}}
\label{LearningRate}
\end{equation}
\begin{equation}
\vspace{-0em}{
\resizebox{.7\linewidth}{!}{$
lr_{top}=\left\{\begin{array}{ll}
lr_{max}\times i/N_{w} & \forall i\in[0,N_{w}]
\\ 
lr_{max}  & \forall i \in (N_{w},N_{a}]
\end{array}\right.$,}}
\label{LearningRate1}
\end{equation}
where $i$ is the iteration index, $N_{w}$ is the number of iterations to warm up the learning rate, $lr_{max}$ is the maximum learning rate which was set to 0.1 for CIFAR and 0.2 for ImageNet-1K, and $N_{a}$ is the total number of iterations. Meanwhile, the weight initialization was used \cite{he2015delving}, and the weight decaies of the optimizers were set to $10^{-4}$ for CIFAR and $10^{-5}$ for ImageNet-1K. Nesterov momentums \cite{sutskever2013importance} of optimizers were 0.9 without dampening. 
\subsection{Experiments on CIFAR}
Generally, the existing DNNs can be categorized into high efficiency and lightweight models, e.g., MobileNets and ShuffleNets \cite{zhang2018shufflenet}, and heavy and complex models for high accuracy, e.g., PyramidNets \cite{han2017deep}. The proposed network could also be scaled into various levels. We designed two Micro-Dense Nets for CIFAR classification, namely Micro-Dense Net 30$\alpha$64 and Micro-Dense Net 60$\alpha$115, and their architectures are shown in Table \ref{cifar_arch}. \\
\indent From Tables \ref{light} and \ref{accuracy on CIFAR dataset2}, we can see that the proposed Micro-Dense Nets, which have the fewest number of parameters, reach the lowest error rates on both CIFAR-10/100. Specifically,  compared with manually designed DNNs, e.g., ResNets and MixedNet, Micro-Dense Nets reach a compatible performance with up to 12.1 $\times$ smaller than state-of-the-art DNNs. Compared with the deep-wise convolution proposed by MobileNets, our cardinality adaptive convolution could further improve the network performance. Moreover, owning to the sufficient utilization of features, the proposed Micro-Dense Nets even achieve compitable performance with the state-of-the-art NAS based method, e.g., FPNASNet.
\begin{table}
	%	\scriptsize
	%	\setlength{\tabcolsep}{0.02\columnwidth}{
	\setlength{\belowcaptionskip}{-0.3cm}
	\resizebox{\linewidth}{!}{
		\begin{tabular}{l r  r  r  r}
			\toprule
			Model						 &\#Params(M)	&Ratio-to-ours	&CIFAR-10				 	&CIFAR-100\\
			\midrule
%			{Micro-Dense Net} 	  		  &0.7			&1.0$\times$	&4.62						&23.61\\
%			{Sparse Net} [ECCV 2018]	 &4.4			&6.3$\times$ 	&3.43						&\color{blue}{19.71}\\
			{GoogLeNet}\protect\cite{szegedy2015going}				 &3.5			&5.0$\times$	&7.06						&27.10\\
			{MobileNet}	\protect\cite{sandler2018mobilenetv2}				  &2.2			&3.1$\times$ 	&4.13						&- \\
			{ShuffleNet} \protect\cite{zhang2018shufflenet}	  &2.5			&3.6$\times$	&5.83						&-	\\
			{ResNet}  \protect\cite{he2016deep} 	 &1.7			&2.4$\times$	&6.43						&-		\\%from dense net
			{ResNet+}	\protect\protect\cite{wang2019convolutional} 				&1.7			&2.4$\times$ 	&6.43						&-		\\
			{Network in network} \protect\cite{lin2013network}		 &1.7			&2.4$\times$	&5.46						&24.33\\
			{FPNASNet} \protect\cite{cui2019fast}		 &1.7			&2.4$\times$ 	&3.99						&-	\\
			{ResNet with Stochastic Depth}[ECCV-2016]&1.7			&2.4$\times$ 			&5.25						&24.98\\
%			{ResNet preactive}			  &1.7			&2.4$\times$ 			&5.46						&24.33	\\
			{PyramidNet($\alpha$=48)} \protect\cite{han2017deep}	  &1.7			&2.4$\times$ 			&4.58						&23.12\\
			{Highway network} \protect\cite{srivastava2015highway}			 &1.7			&2.3$\times$	&7.54						&-\\
			{MixNet} \protect\cite{wang2018mixed}       &1.5			&2.1$\times$	&4.19						&21.12 \\
			{DenseNet} \protect\cite{huang2017densely}		  &1.0			&1.4$\times$ 			&5.24						&24.33\\
			{Micro-Dense Net-30$\alpha$64} 	  	  &0.7			&1.0$\times$	&\color{blue}{3.41}			&\color{blue}{20.85}\\
			{Micro-Dense Net-30$\alpha$64+}   	  &0.7			&1.0$\times$	&\color{red}{3.28}			&\color{red}{19.47}\\
			\bottomrule
	\end{tabular}}
	\vspace*{-0.9em}
	\caption{Error rate (\%) of different lightweight methods. '+' denotes the method using dynamic regularization \protect\cite{wang2019convolutional}. The lowest (the best) and second lowest error rates are highlighted with red and blue, respectively.}
	\label{light}
\end{table}
\begin{table}
	\scriptsize
	\setlength{\belowcaptionskip}{-0.5cm}
	\resizebox{\linewidth}{!}{
		%	\begin{tabular}{m{0.38\columnwidth} m{0.1\columnwidth}  m{0.1\columnwidth}  m{0.09\columnwidth}  m{0.09\columnwidth}}
		\begin{tabular}{lrrrr}
			\toprule
			Model						 			&\#Params(M)   	&Ratio-to-ours  	&CIFAR-10				 		&CIFAR-100	\\
			\midrule
			{MixNet} \protect\cite{wang2018mixed} 		  			&48.5			&12.1$\times$  			&3.13							&16.96 \\
			{Wide ResNet} \protect\cite{cubuk2019autoaugment}			 	&36.5			&9.1$\times$ 			&2.6							&17.1\\
			{Res2NeXt-29} \protect\cite{gao2019res2net}				&36.7			&9.2$\times$  			&-								&16.79\\
			{Res2NeXt-29-SE} \protect\cite{gao2019res2net} 			&36.9			&9.2$\times$  			&-								&\color{blue}{16.56}\\
			%		{SEnet} [CVPR-2018] 			  		&				&34.4				&\color{red}{2.12}				&\color{blue}{15.41}\\
			{PyramidNet($\alpha$=270)} \protect\cite{han2017deep}	&27.0			&6.8$\times$ 			&3.48							&17.01\\
			{DenseNet} \protect\cite{huang2017densely}		  			&25.6			&6.4$\times$  			&3.46							&17.18\\
			{SparseNet} \protect\cite{zhu2018sparsely}  			&16.7			&4.2$\times$  			&3.22							&17.71\\
			{DenseNet} \protect\cite{huang2017densely}		  			&15.3			&3.8$\times$ 			&3.62							&17.60\\
			{Wide ResNet} \protect\cite{tan2019mnasnet}			  	&11.0			&2.8$\times$ 			&4.27							&20.43\\
			{FPNASNet} \protect\cite{cui2019fast}  		  		&5.7			&1.4$\times$			&3.01							&-	\\
			{NnasNet-A} \protect\cite{zoph2018learning}  		  		&3.3			&0.8$\times$			&2.65							&-	\\
			{Micro-Dense Net-60$\alpha$115}	  		&4.0			&1.0$\times$				&\color{blue}{2.58}				&16.92\\
			{Micro-Dense Net-60$\alpha$115+}   		&4.0			&1.0$\times$				&\color{red}{2.16}				&\color{red}{15.36}\\
			\bottomrule
	\end{tabular}}
	\vspace*{-0.8em}
	\caption{Error rate (\%) of different heavy models. '+' denotes the method using dynamic regularization. The lowest and second lowest error rates are highlighted with red and blue, respectively.}
	\label{accuracy on CIFAR dataset2}
\end{table}
\setcounter{figure}{3}
\begin{figure*}
\centering
\setlength{\belowcaptionskip}{-0.5cm}
\includegraphics[width=5cm]{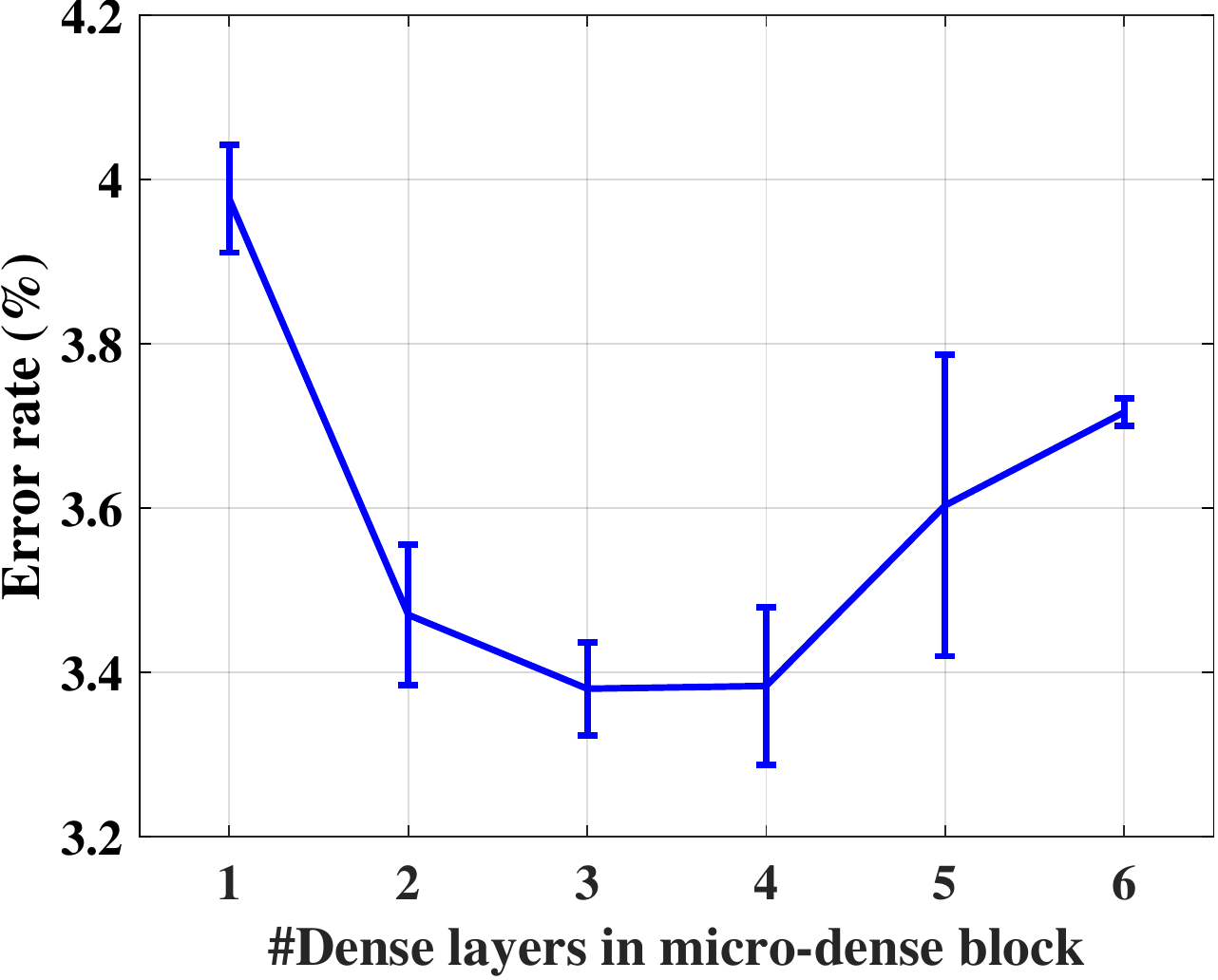}
\includegraphics[width=5cm]{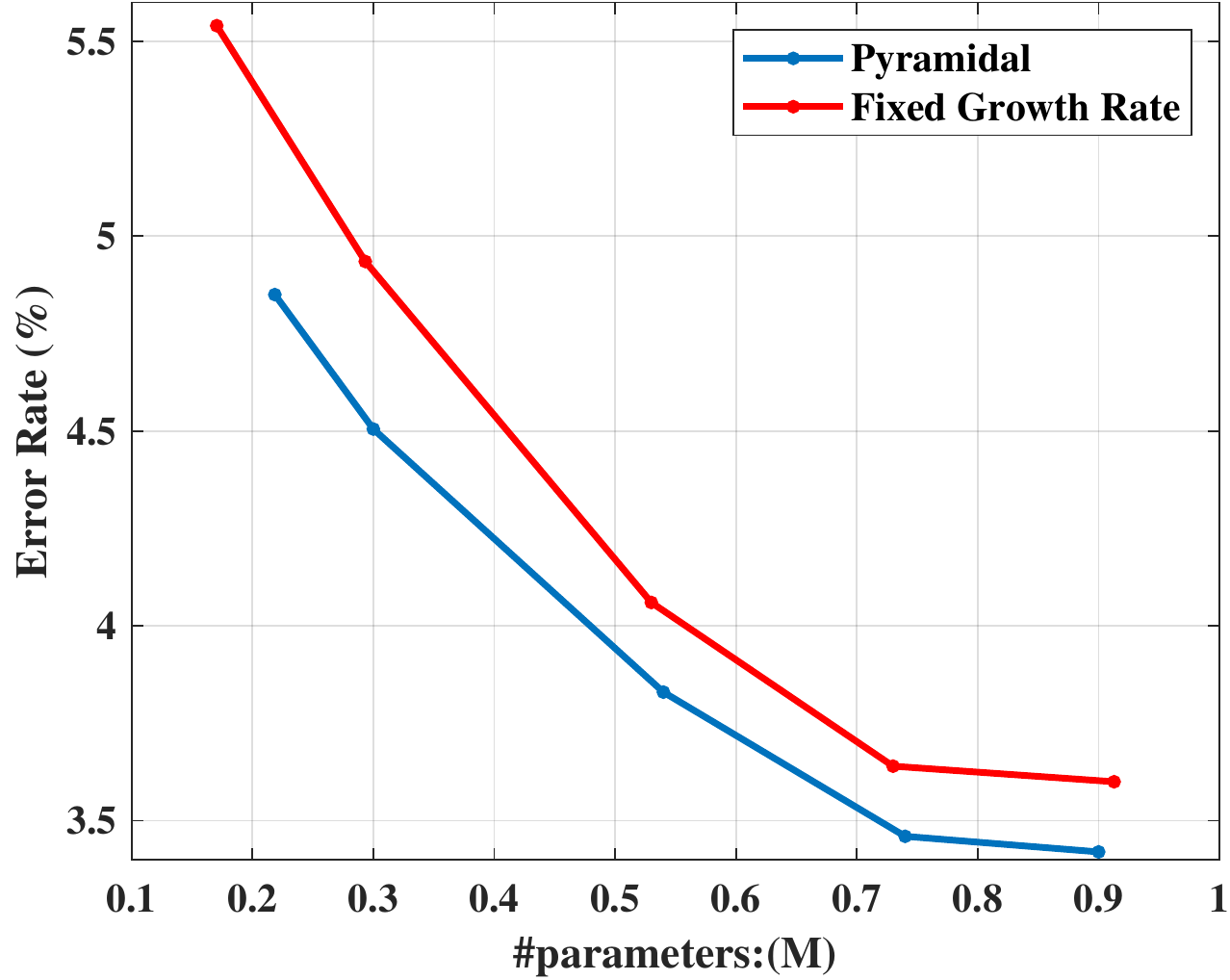}
\includegraphics[width=5.5cm]{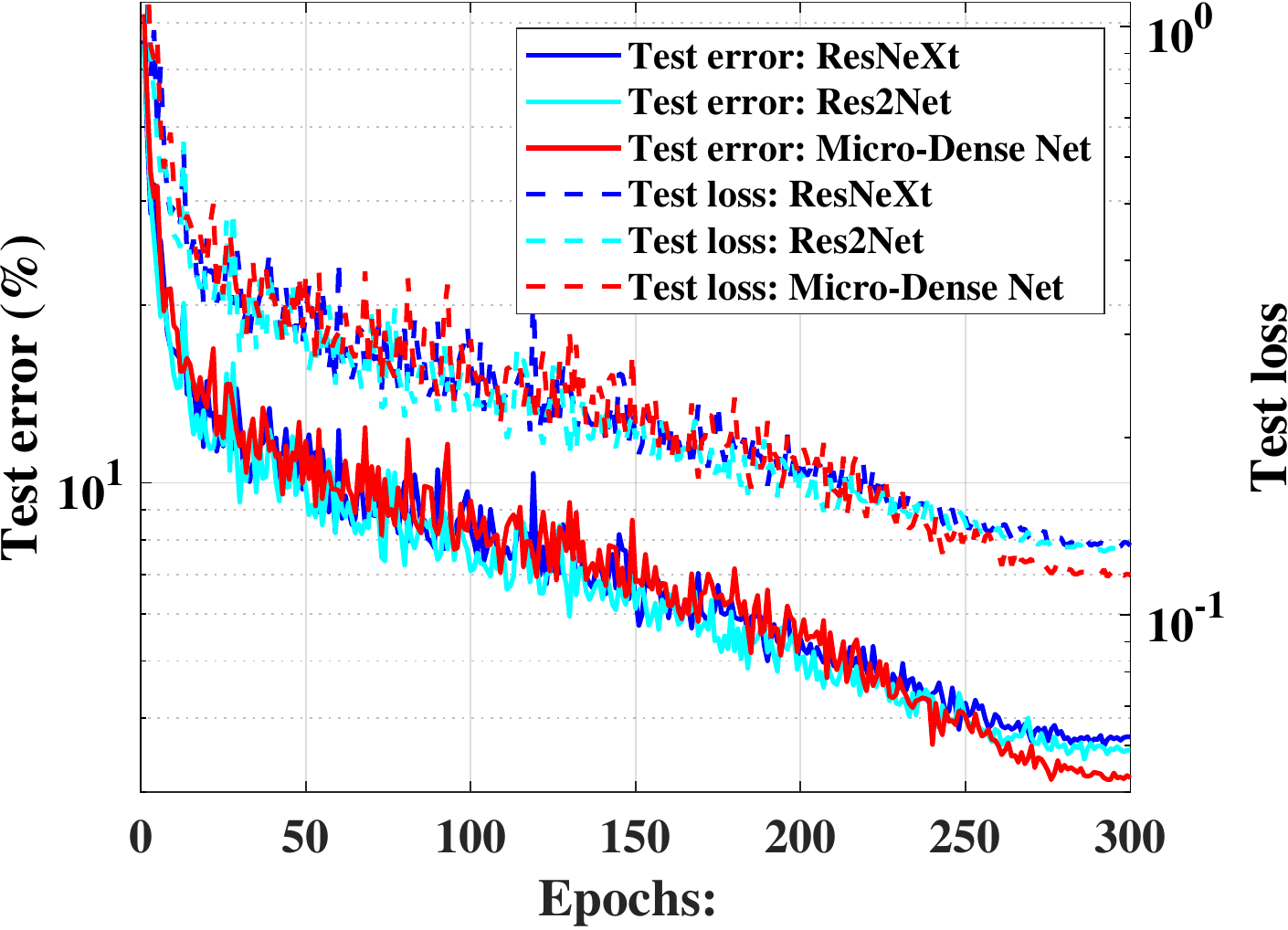}
\vspace*{-0.8em}
\caption{ \textbf{\textsl{Left}}: Error rates of networks with 0.7M parameters but varies with different numbers of dense layers. \textbf{\textsl{Middle}}: Error rates of network with (fixed number / gradually increasing) channels. \textbf{\textsl{Right}}: Error rate of the training process with different networks. We carried out all these experiments on CIFAR-10.}
\label{final result}
\end{figure*}
\subsection{Experiments on ImageNet-1K}
NAS has greatly boosted the performance of DNNs on ImageNet-1K. However, due to the limitation of the searching space, it is still hard for NAS to find the optimal neural network topology. Here, we integrated the proposed micro-dense architecture with the state-of-the-art classification model EfficientNet-B0 \cite{tan2019efficientnet}. Table \ref{imagenet result} shows the top-1 and top-5 test accuracies on the ImageNet-1K dataset. We replaced the $8$-th,$9$-th,$10$-th \textsl{MBConv} modules in EfficientNet-B0 with our micro-dense block under the same number of parameters and FLOPs, and the resulting new model is denoted by Micro-Dense-EfficientNet. It is also worth noting that the other part of Micro-Dense-EfficientNet are the same as the original EfficientNet-B0 one which are not optimized for \textsl{MDConv-n}. The \textsl{MBConv} blocks of EfficientNet-B0 are obtained from NAS algorithm \cite{tan2019mnasnet}.\\
\begin{table}
	\centering
	\setlength{\belowcaptionskip}{-0.4cm}
	\resizebox{\linewidth}{!}{
		\begin{tabular}{l r r r r }
			\toprule
			Model						 &Params(M) 		& FLOPs(B)		&Top-1 Acc.			&Top-5 Acc.			\\
			\midrule
			{ResNet-50} 		  		  &26.0				&4.1		&76.0				&93.0 				\\
			{DenseNet-169}    			  &14.0				&3.5		&76.2				&93.2				\\
			%			{MnasNet-A2}	  	          &4.8				&75.6		    	&92.6				&4.1B\\	
			{NASNet-A (4 @ 1056)}		  &5.3				&0.6		&74.0				&91.6				\\
			{EfficientNet-B0}			  &5.3				&0.4		&76.3				&93.2 				\\
			{EfficientNet-B0*}			  &5.3				&0.4		&76.5				&93.0 				\\
			{Micro-Dense-EfficientNet}	  &5.3				&0.4		&\color{red}{76.8}	&\color{red}{93.2}	\\		  
			\bottomrule
	\end{tabular}}
	\vspace*{-0.6em}
	\caption{Accuracy (\%) of state-of-the-art ImageNet-1K classification models including both the manually designed and NAS based methods. '*' denotes our implementation.}
	\label{imagenet result}
\end{table}
\indent As listed in Table \ref{imagenet result}, experimental results reveal that our Micro-Dense-EfficientNet could also perform on par with NAS methods. That is, Micro-Dense-EfficientNet boosts the performance of EfficientNet-B0 by 0.5\% point on Top-1 accuracy, which validates the advantage of our architecture. The activation mappings (thorugh \cite{selvaraju2017grad}) in Figure \ref{final result_0} also show that micro-dense architecture could help DNNs to get more accurate concentrations on the targets. Moreover, it is expected that replacing searching space from \textsl{MBcov} to \textsl{MDConv-n} could further rise up the frontier of feature learning.
\subsection{Architecture Analysis}
\setcounter{figure}{2}
\begin{figure}
	\Large
	\setlength{\belowcaptionskip}{-0.5cm}
	\resizebox{\linewidth}{!}{
		\begin{tabular}{c ccccc}
			\toprule
			Plough &Hockey Puck & Poncho &  Wagon & Missile\\
			\midrule
			\rotatebox{90}{ \qquad Original} \includegraphics[width=3cm]{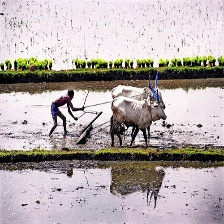}& \includegraphics[width=3cm]{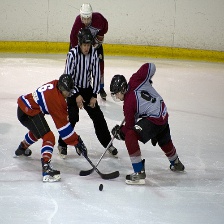} & \includegraphics[width=3cm]{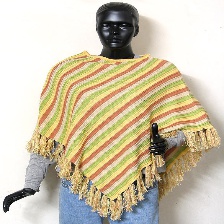}  &\includegraphics[width=3cm]{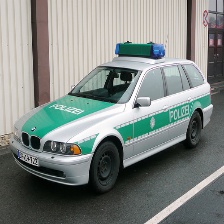} &\includegraphics[width=3cm]{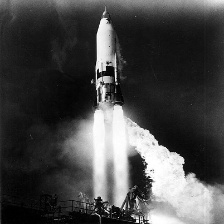} \\
			\rotatebox{90}{ \quad EfficientNet} \includegraphics[width=3cm]{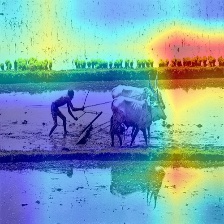}&  \includegraphics[width=3cm]{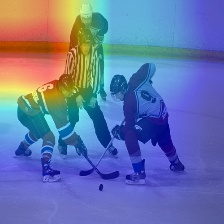} & \includegraphics[width=3cm]{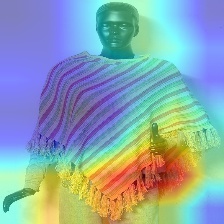}  &\includegraphics[width=3cm]{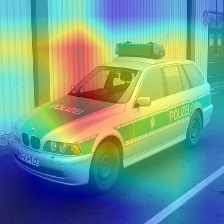} &\includegraphics[width=3cm]{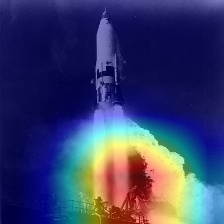} \\
			\rotatebox{90}{ \qquad Ours} \includegraphics[width=3cm]{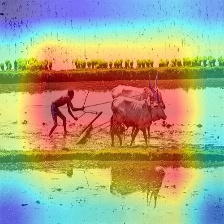} &  \includegraphics[width=3cm]{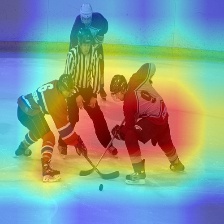} & \includegraphics[width=3cm]{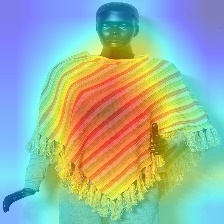} & \includegraphics[width=3cm]{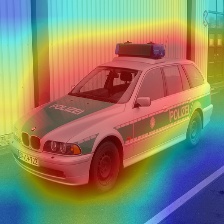} &\includegraphics[width=3cm]{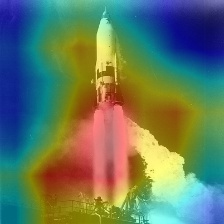} \\
			\bottomrule
		\end{tabular}
	}
	\vspace*{-0.9em}
	\caption{\textsl{\textbf{Top}}: input images. \textsl{\textbf{Middle}}: activation mappings from EfficientNet.  \textsl{\textbf{Bottom}}: activation mappings from Micro-Dense-EfficientNet. They represent the attentions of DNNs.}
	\label{final result_0}
\end{figure}
\indent We carry out more experiments to investigate the proposed micro-dense architecture. \\
\textbf{Micro-dense vs. dense aggregations}. The dense aggregation could fully exploit features
from different levels, but it also consume much resource on preserving and fusing accumulated feature-maps from proceeding layers. 
Thus, our micro-dense blocks with different
branches but same number of parameters and depths are evaluated in this section. The result is shown in the left of Figure \ref{final result}. We can observe that:
1) the local dense aggregation indeed boosts the performance of DNNs as the error rate declines with the number of dense layers increasing at the beginning stage; and
2) very deep dense aggregation is not necessary for the micro-dense block, and it may even degrade the performance of DNNs. \\
\begin{table}
	\centering
	\scriptsize
	\setlength{\belowcaptionskip}{-0.6cm}
	\begin{tabular}{l r r r r }
		\toprule
		Model				 			&CIFAR-10			&CIFAR-100		\\
		\midrule
		{ResNeXt}						&3.70				&20.95 			\\
		{Res2Net}						&3.49 				&20.45 		\\
		{Micro-Dense Net}				&\color{red}{3.35}	&\color{red}{18.49}	\\		  
		\bottomrule
	\end{tabular}
	\vspace*{-0.8em}
	\caption{Average error rate (\%) of state-of-the-art models: Res2Net, ResNeXt with Micro-Dense Net. All the methods are with same number of parameters 1.8M and FLOPs 0.26G. The only variable is architecture of DNNs.}
	\label{Dense_aggregation_result}
\end{table}
\textbf{Pyramidal vs. fixed growth rate multi-level feature learning.} In the micro-dense block, pyramidal multi-level feature learning with varying dimension cardinality is proposed to extract features efficiently. To validate pyramidal multi-level feature learning, We compare Micro-Dense Nets with ones trained with a fixed growth rate. The experimental results in the middle of Figure \ref{final result} shows that pyramidal multi-level feature learning strengthens the representation learning ability of DNNs. \\
\textbf{Micro-dense vs. residual aggregations.} To investigate the effectiveness of micro-dense aggregation in residual learning, we compare Micro-Dense Nets with state-of-the-art residual learning based methods Res2Next and Res2Net with same number of parameters and FLOPs. Note we only replaced each ResNeXt block with Res2Net or micro-dense block and they were trained under exact same conditions. From Table \ref{Dense_aggregation_result}, we can see that Micro-Dense outperforms Res2Next and Res2Net, indicating the feature learning ability is boosted. Meanwhile, the test errors and losses are also compared in the right of Figure \ref{final result}, and our Micro-Dense Net achieve the lowest errors and loss on both CIFAR-10/100.\\ 
\section{Conclusion}
    In this paper, we proposed a new DNN architecture, namely Micro-Dense Nets, which takes advantages of identity mapping and dense aggregation to fuse features from different levels. Moreover, Micro-Dense Nets are characterized as pyramidal multi-level feature learning and dimension cardinality adaptive convolution. Extensive experimental results demonstrate that Micro-Dense Nets achieve state-of-the-art performance over CIFAR-10/100 and ImageNet-1K. In addition, Micro-Dense Nets require much fewer parameters and computational resources than ResNets and DenseNets. Last but not least, Micro-Dense Net could also be integrated with NAS based models to boost their performance, which forcefully validates the advantage of our design.
\bibliographystyle{named}
\bibliography{main}
\end{document}